# Simulating multi-exit evacuation using deep reinforcement learning

Dong Xu[a], Xiao Huang[b], Joseph Mango[a], Xiang Li[a] and Zhenlong Li[b]

[a]Key Laboratory of Geographical Information Science (Ministry of Education), School of Geographic Sciences, East China Normal University, Shanghai, China

[b]Geoinformation and Big Data Research Laboratory, Department of Geography, University of South Carolina, SC, USA

Conventional simulations on multi-exit indoor evacuation focus primarily on how to determine a reasonable exit based on numerous factors in a changing environment. Results commonly include some congested and other under-utilized exits, especially with massive pedestrians. We propose a multi-exit evacuation simulation based on Deep Reinforcement Learning (DRL), referred to as the MultiExit-DRL, which involves in a Deep Neural Network (DNN) framework to facilitate state-to-action mapping. The DNN framework applies Rainbow Deep Q-Network (DQN), a DRL algorithm that integrates several advanced DQN methods, to improve data utilization and algorithm stability, and further divides the action space into eight isometric directions for possible pedestrian choices. We compare MultiExit-DRL with two conventional multi-exit evacuation simulation models in three separate scenarios: 1) varying pedestrian distribution ratios, 2) varying exit width ratios, and 3) varying open schedules for an exit. The results show that MultiExit-DRL presents great learning efficiency while reducing the total number of evacuation frames in all designed experiments. In addition, the integration of DRL allows pedestrians to explore other potential exits and helps determine optimal directions, leading to a high efficiency of exit utilization.

## 1. Introduction

Indoor pedestrian simulation in evacuation studies is one of the most critical components that has been receiving great attention due to its potential of rescuing people in any case of emergency (Sun and Li 2011, Chen and Feng 2009). A reasonable simulation should be able to guide the pedestrian to leave fast and safely through exits of the indoor environments. In order to achieve this objective, pedestrians need to choose the best route to pass through, usually depending on the number of sub-exits present in a particular complex. This problem can be regarded as a navigation planning problem where the shortest path can be therefore chosen by using sampling-based algorithms (Kuffner and LaValle 2000, Karaman and Frazzoli 2011) or geometry-based algorithms (Geraerts 2010, Kallmann 2014). The pedestrians are expected to reach the final exits in the shortest time as long as they have followed the calculated path. However, these assumptions are only effective and efficient in the environment where the number of pedestrians and sub-exits is low. In actual sense, there many complex indoor environments with a large number of pedestrians. In such circumstances, the evacuation process becomes more complex due to overcrowding, starting from the nearest sub-exits to the final exits. This overcrowding problem is usually due to the narrow widths of the sub-exits, which pose challenges for pedestrians to leave in a rapid manner. Moreover, this problem becomes more complex if there is a variation of the sub-exit widths with uneven distribution of the pedestrian. Therefore, how to reasonably allocate pedestrians to different exits is still a valid and challenging question in pedestrian evacuation simulation studies.

The existing multi-exit selection research can be broken into two categories of studies. The first category is focused on the building design, aiming at proposing a more reasonable and more efficient multi-exit layouts (Seyfried et al. 2009, Choi et al. 2014). However, the models designed from these studies are unable to accommodate pedestrian behaviors that are proven to be significant during the evacuation process (Bode and Codling 2013). The second is focused on designing a multi-exit selection model using realistic experimental simulation (Haghani et al. 2015, Heliövaara et al. 2012, Wagoum et al. 2017). As contrary to the first category, the models designed using the second approach often require high preliminary costs. Diverging from these two main streams, other researchers started to utilize virtual simulation to describe the multi-exit simulation process of pedestrians by setting motion in models to guide the pedestrian navigation via movement strategies. Following this direction, Zia and Ferscha (2009) proposed three different strategies to analyze the evacuation efficiency of pedestrians in the room. In the same domain, Hao et al. (2014) also proposed a mix-strategy to combine distance-based and time-based strategy based on an improved dynamic parameter model.

In the past decade, deep learning as one branch of the machine learning methods emerged to provide remarkable modeling results such as in image classification (Krizhevsky et al. 2012), object tracking (Bertinetto et al. 2016), and natural language processing (Sutskever et al. 2014). As one type of deep learning, Deep Reinforcement Learning (DRL) has also made significant successes in defeating human players in many competitions such as in Go (Silver et al. 2016) and Arita games (Mnih et al. 2013). Unlike supervised learning and unsupervised learning, DRL learns a mapping (i.e., from state to action) to maximize the long-term reward (Sutton and Barto 2018). Depending on its learning methods, DRL can be classified into two categories of Deep Q-Network (DQN) and Policy Gradient (PG). To date, there are several advanced DRL algorithms which include: Rainbow DQN (Hessel et al. 2018), Proximal Policy Optimization (PPO) (Schulman et al. 2017), Soft Actor Critic (SAC) (Haarnoja et al. 2018), and Deep Deterministic Policy Gradient (DDPG) (Lillicrap et al. 2015). All these DRL algorithms have been applied in different areas, such as in robot visual navigation. Typical examples in this area of application are found in some studies such as the study by Gupta et al. (2017) that proposed a cognitive mapper and planner based on a neural architecture to navigate the robot using the first-person views as the state space. Another study by Zhang et al. (2017) presented the application of DRL-based robot navigation in a maze-like environment, where the robot has shown a robust adaptive ability for new environments. Apart from the examples of robotic applications, DRL algorithms have been also used in local motion planning as well. In this case for example, Lee et al. (2018) proposed a deep reinforcement learning method based on the AC framework for crowd simulation. In the same field, Long et al. (2018) presented a decentralized collision avoidance policy by directly detecting the environment using raw sensors and feeding the measurements to a DRL network. Despite these achievements, the research on multi-exit navigation based on DRL needs to be explored further in different environments.

In this paper, we focus on exploring the potential of DRL in multi-exit navigation within certain designed room environments. Each pedestrian is defined as a disc, characterized by its position and velocity. The simulation process in our research can be regarded as a series of ordered frames. At a given frame, the pedestrians' positions are determined based on their velocities (described in detail in Section 3). A total of eight possible movement directions are provided for each pedestrian, and the desired direction is determined using the DRL algorithm. The Optimal Reciprocal Collision Avoidance (ORCA) is adopted to avoid collision during the pedestrians' movement. Unlike other studies, we do not explicitly assign a reasonable exit. In our method, each pedestrian is considered as an intelligent person with self-judgment, who takes desired actions to facilitate its evacuation after millions of iterations of interacting with the environment. To illustrate the advantages of our method, we compared our proposed method with two popular methods from (Zheng et al. 2015) and (Guo et al. 2012). The main contributions of this paper are threefold: 1) we designed a hierarchical model to handle multi-exit navigation simulation in micro-scale, a more realistic method in describing pedestrian's behaviors compared with other simulations in mesoscale; 2) we developed a new method which abstracts the grey image of the room as the state space by distinguishing the target pedestrian and other pedestrians using greyscale values. Compared with the traditional ray casting methods that consider external environments (Lee et al. 2018), our method can noticeably speed up the simulation process; 3) we integrated Rainbow DQN, an effective DRL method that combines several advanced DQN algorithms. This training design largely increases the stability of the neural network and the speed to reach the convergence point. Furthermore, the Rainbow DQN in our method can be updated in One-Step or N-Steps without collecting a complete training trajectory. Thus, it provides a privilege in data collection compared with the traditional PG methods.

This work is organized into seven sections, including this part of the introduction. Section 2 summarizes the existing literature of multi-exit navigation and deep reinforcement learning algorithms. Section 3 presents more information about ORCA and the Rainbow DQN algorithm in network architecture, state-space, action-space, and reward functions. Section 4 introduces three designed scenarios to compare our proposed MultiExit-DRL model with two traditional models based on mathematical strategy navigation. It also describes the settings of the relevant parameters and computer configurations used when developing our model. Section 5 presents, analyzes, and discusses the results of the proposed MutliExit-DRL model. Section 6 discusses the advantages and limitations of the proposed model before concluding our work in Section 7.

## 2. Related Work

### 2.1 Pedestrian simulation

Nowadays, Pedestrian simulation has received large attention in geographic information science (GIS) field and widely used in path planning (Wu et al. 2007), emergency decision making (Tashakkori et al. 2015), and human behavior analysis (Li et al. 2010). Pedestrian simulation can be divided into two groups depending on their simulation scales: these are, the meso-scale simulation and the micro-scale simulation. Cellular Automata (CA) model, as a classic meso-scale model, has been applied to simulate the movement of pedestrians and reached decent results (Chopard and Droz 1998). In general, a CA model divides the research region into a series of regular cells, in which each of them holds different statuses, including "available", "occupied", and "obstacle". A pedestrian can move to the adjacent cells in the next frame if there are available cells surrounding the current cell. In the past decades, the advancement of the computational power of rough modeling facilitated many CA related studies (Dijkstra et al. 2001, Burstedde et al. 2001, Pelechano and Malkawi 2008). However, the disadvantages of the CA model are also known due to the limitation of rough modeling. In particular, the CA model discretizes pedestrian movements, falling short of describing the dynamic behaviors of pedestrians (Cao et al. 2016).

In light of this issue, micro-scale pedestrian simulation has gradually become one of the research hotspots. Unlike the rough modeling, commonly used in meso-scale simulations, microscopic models accurately simulate the environment and pedestrians with finer details via geometric expressions, thus to allow pedestrians to move arbitrarily within the configuration space. The aforementioned advantage made micro-scale simulation more popular and has attracted many types of research in different directions. For instance, Helbing and Molnar (1995) proposed a classic social force model (SFM), which characterizes the interactions between pedestrians using a mathematical model with pedestrian behaviors. The SFM has received widespread attention due to its convenient and effective nature (Mehran et al. 2009, Hou et al. 2014, Karamouzas et al. 2017). Nevertheless, SFM and other force-based methods fail to guarantee complete collision avoidance (Curtis and Manocha 2014). In comparison, geometrical based methods are considered to be better methods to avoid collisions and have been widely applied in robotics and motion simulation (Curtis and Manocha, 2014). Velocity Obstacle (VO) method is one of the traditional geometry-based collision avoidance method which is able to forecast potential collisions by calculating the VO region (Fiorini and Shiller 1998). As an improvement of VO approach, a reciprocal n-body collision avoidance approach (ORCA) by Van Den Berg et al. (2011) introduced a low-dimensional linear program for collision-free movements. As reviewed in many other pieces of literature, the ORCA approach proved to simulate collision-free actions for thousands of agents in a few milliseconds. Due to its efficiency, the ORCA has been applied in many applications with different modifications. For instance, Alonso-Mora et al. (2013) guaranteed a smooth and collision-free motion for non-holonomic robots via distributed collision avoidance simulation under non-holonomic constraints. Golas et al. (2013) presented an effective algorithm to perform long-range collision avoidance in crowd simulation by adopting a novel metric to quantify the smoothness of trajectories. Bareiss and van den Berg (2015) proposed a generalized reciprocal collision avoidance algorithm, not only for presenting an extension to control obstacles but also for generating collision-free motions under a non-linear and non-homogeneous system. Given the superiorities of ORCA method and its flexibility in different applications, we adopted it in our study as the local pedestrian simulation method to avoid collision during the simulation of the multi-exit evacuation process.

**2.2 Multi-exit selection**

Evacuation simulation has always been a hot topic in the fields of safety and robotics. Compared with common pedestrian simulations, evacuation simulation focuses more on the pedestrian behaviors given in various internal or external factors during the evacuating process. For example, Parisi and Dorso (2005) applied SFM to explore the 'faster is slower' effect under different degrees of panic or fear. Frank and Dorso (2011) investigated the efficiency of evacuation by placing some obstacles near the exit. Ben et al. (2013) presented an agent-based modeling approach to describe individual behavior during the evacuation. Wang et al. (2015a) developed a novel multi-agent based congestion evacuation model to investigate individual panic behavior at the individual level and further analyze the evacuation efficiency if a virtual leader is added in the model.

As an important component in evacuation simulation, multi-exit selection aims to mimic or analyze pedestrians' behaviors under complex local environments. In some studies, the evacuation data are extracted from realistic scenarios, and the decision rules are specified accordingly (Guo et al. 2012, Haghani and Sarvi 2016). Such approaches can model pedestrian behavior in an accurate manner, however, it is difficult to archive since the realization of the realistic scenarios is always elusive. Alternatively, other studies focus on investigating the evacuation process via computer simulations by controlling the designed environments (Bode and Codling 2013, Kinateder et al. 2014, Davidich et al. 2013, Han et al. 2017, Zhou et al. 2019). Nonetheless, the choice of exits in this direction also remains to be a great challenge, especially when dealing with complex environments (Zheng et al. 2017, Zia and Ferscha 2009).

In most studies, the choices of exits are determined by several factors, including the distance from pedestrians to the exits, the width of exits, the degree of congestion, and the familiarity of pedestrians with the environment. Fu et al. (2018) proposed an exit selection method during evacuation processes based on the CA mode; it considers the impact of the building design on the pedestrians' behaviors. To investigate the multi-exit selection, Wagoum et al. (2017) conducted three empirical experiments with several extracted indicators that link temporal information with the choice of exit behaviors. Kinateder et al. (2018) developed an exit selection model in an ambulatory virtual environment, aiming to reveal the 'movement to the familiar' behavior in the controlled experiments. Lo et al. (2006) proposed a novel dynamic exit selection process based on a non-cooperative game theory that describes the equilibrium between the number of evacuees and the number of exits. To investigate the influence of the asymmetry of exits on pedestrian behaviors, Hao et al. (2014) proposed a mixed strategy that fused the distance-based and time-based strategies using a cognitive coefficient. (Cao et al. 2018) described the exit selection based on a random utility theory in a room of two-exits. In addition, other studies applied more complex numerical operations to the multi-exit selection and achieved remarkable results. For example, Guo et al. (2012) proposed a nested logit discrete model (NLDM) to investigate the choices of exits considering different factors such as visibility, intimacy, and physical conditions of the exits. (Zheng et al. 2015) proposed an improved adaptive multi-factor model (AMFM) by considering the factors of spatial distance, density, and the exit width.

## 2.3 Deep Reinforcement Learning (DRL) and its applications in pedestrian simulation

Reinforcement Learning (RL) has attracted wide attention, largely due to its powerful performance in playing games like Go and Arita games (Silver et al. 2017, Mnih et al. 2013). Generally, RL consists of three essential parts: environment, agent, and the reward (Sutton and Barto 2018). An agent takes action based on the current state in the environment, then the environment proceeds to the next state and renders a corresponding reward, serving as an evaluation of the action taken by the agent. Further, the agent takes the next action based on the next state and receives the next reward with a newly updated state. This iteration continues until the stopping rules are reached. The purpose of RL is to discover a function, i.e., a state-to-action mapping, in order to increase the total reward under the current state. Unlike other machine learning methods,

RL focuses on long-term rewards. That's why it is popular in many fields, including complicated decision-making, robot simulation, and intellectual games.

In recent years, many studies have been done to combine deep neural networks (DNN) and RL to come up with the so-called Deep Reinforcement Learning (DRL) for solving more complicated problems. For instance, Mnih et al. (2013) proposed a novel DRL method, named Deep Q Network (DQN), by combining Q-Learning (an important branch in RL methods) with DNN. The results showed that it was able to achieve excellent results that surpassed the human-level performances in multiple Atari games. To reduce the problem of overestimation in DQN, Double DQN was developed by decoupling the target value (Van Hasselt et al. 2016). Instead of generating action value directly using a fully connected network, Dueling DQN (DDQN) first decouples the action value in DNN with the state and advantage values and then, it combines the two values as a total action value (Wang et al. 2015b). Benefitting from this network architecture, DRL can easily learn the importance of the states.

On the other hand, there are researches that explored the learning efficiency of the DRL based on the sampling method. For example, Schaul et al. (2015) proposed a Prioritized Experience Replay (PER) algorithm that orders data according to the loss value, and in the end, the algorithm proved to have higher sample utilization. Instead of using epsilon greedy to balance the tradeoff between exploration and exploitation, Noisy DQN is proposed to encourage exploration by generating random noises in the DNN (Fortunato et al. 2017). The result showed that Noisy DQN could yield better scores in Atari games as it leads to more efficient exploration. To make the learning process more stable, Bellemare et al. (2017) proposed a Categorical DQN algorithm by approximating value distribution using expectation value. The result showed that Categorical DQN facilitates chattering reducing, state aliasing, and well-behaved optimization. Recently, Hessel et al. (2018) proposed a Rainbow DQN approach that integrates the several independent improvements of the DQN algorithm. Compared with other algorithms, Rainbow DQN has noticeable improvements in the reward and convergence speed. Given its great capability, Rainbow DQN has been widely applied in open car simulation (Güçkıran and Bolat 2019), adaptive traffic signals (Nawar et al. 2019), and predictive panoramic video streaming (Xiao et al. 2019).

Nowadays, DRL is extensively used for robot navigation and pedestrian simulation. In robot navigation, the robots are able to detect the environment with the support of raw cameras or other types of sensors. In this case, the DRL algorithm is responsible for generating an optimal action to navigate the robot to the destination without collisions with other obstacles. For example, Zhang et al. (2017) applied successor-feature-based DRL in robot navigation with maze-like environments. The results suggested that this algorithm can easily adapt to other new environments. Kahn et al. (2018) proposed a self-supervised DRL with a generalized computation graph to autonomously navigate the robot in real-world environments. It finally proved to have high sample efficiency in learning complex policies. For the crowd simulation, virtual pedestrians are the agents in the environment (usually a micro-scene), aiming to reach their destinations. During the moving process, each pedestrian should avoid static and dynamic obstacles and ensure the trajectory without noticeable oscillations. Numerous attempts have been made to use DRL in crowd simulation. Godoy et al. (2016) proposed a novel Coordinated Navigation (C-Nav), a distributed approach, to address the multi-agent navigation problem in complex environments. The result showed great generalization, and the agents can reach their goals faster than other advanced collision-avoidance frameworks. Long et al. (2018) proposed a decentralized collision avoidance policy based on DRL for multi-robot systems where the inputs are directly detected from the raw sensors. To smoothen the global trajectory of agents, Xu et al. (2020) presented a local motion simulation method that integrates ORCA and DRL, named as ORCA-DRL. The model has shown a great capability of generalization, smoothening the global trajectory, and fastening learning, compared with other DRL methods. Nevertheless, the proximal policy optimization (PPO) algorithm applied in Xu et al. (2020) falls short to adapt a dynamic number of agents. In addition, the state inputs based on the traditional raycasting method largely limit its performance. Despite the aforementioned advancements, the exploration of the DRL's potential in multi-exit evacuation simulation by involving room environments is still rare.

In order to fill the gap, we proposed a novel multi-exit evacuation simulation, named as MultiExit-DRL. In this prototype, the ORCA is applied to avoid collision of pedestrians during the simulation, and Rainbow DQN architecture is used for guiding pedestrians to choose the best direction. Unlike in other studies of multi-exit evacuation simulation, this study does not explicitly define the best choice of exits for the pedestrian to evacuate at each frame. Instead, we defined eight directions in each frame for them to learn the best choice using Rainbow DQN. We further designed several indoor environments to illustrate the advantages of integrating DRL in multi-exit environments. The designed MultiExit-DRL model has demonstrated good results after comparing it with the other two traditional models, and therefore, it can be applied in many local indoor environments to investigate the behavior of the pedestrian flows. It can also be applied as the component to evaluate the evacuation efficiency under different building designs.

## 3. Methods

### 3.1 Methodology overview

We set the simulation environment as a two-dimensional space $\mathbb{R}^2$, which consists of $N_{ped}$ randomly distributed pedestrians and $N_{exits}$ exits with varying properties. The goal of each pedestrian is to evacuate the room as quickly as possible through one of the available exits without colliding with obstacles or other pedestrians. Each pedestrian is represented by a disc of the radius $r$ with the maximum speed of $v_i^{max}$. The obstacles are composed of line segments defined by a series of counter-clockwise points. The pedestrian $i$ in frame $t$ is characterized by the following parameters:

- position as $p_i^t$
- velocity as $v_i^t$
- speed as $v_i^t$
- collision-free velocity, i.e., optimal velocity as $^{opt}v_i^t$
- direction as $a_i^t$

Each exit holds two properties. For exit $j$, $\boldsymbol{p}_{exit_j}$ represents the position of exit $j$ and $w_{exit_j}$ represents the width of exit $j$. The pedestrians in the room are able to observe the positions and velocities of other pedestrians without explicit communication. At each frame within a cycle of a total of $m$ frames, pedestrians update their positions based on the $^{opt}v$ restrained from the kinetics rules (Figure 1). The simulation process continues until all pedestrians successfully evacuated the room, or the number of frames reaches the horizon $T$. At each frame $t$, pedestrians interact with the environment and the resulted interaction data, including state, action, reward, and terminal information, are stored in the container with a capacity of $N_{buffer}$ (Figure 1a). We further define an integer value named, learning start $(L_s)$, to encourage the pedestrian to explore the environment sufficiently. At each training step, we sample interaction data from the container with a defined batch size. DNN parameters are further updated by Rainbow DQN, where a crossing-entropy loss is implemented to minimize the KL divergence between predicted state-action probabilities and target state-action probabilities. At the end of each training step, the priorities are updated based on the loss weight. The interacting process (Figure 1b) at each frame follows the procedure described in Xu et al. (2020). More details regarding the simulation implementation are presented in the following sections.

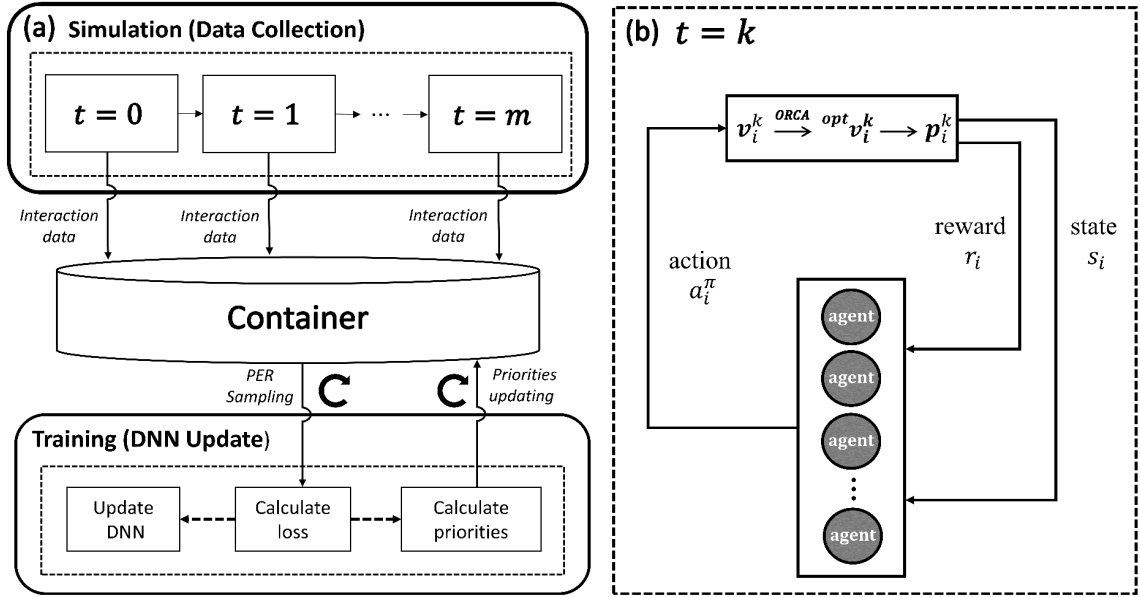

**Figure 1** Methodology overview. (a) General simulation structure; (b) Interacting process for all agents at frame $t = k$ (modified from Xu et al. (2020))

### 3.2 Deep reinforcement learning

#### 3.2.1 State space, action space, and reward function

##### 3.2.1.1 State space

We assume that pedestrians are fully aware of the environment. Similar to the state space in the Arita game, we directly abstract the images in the last three frames as the current state. The image is resized to $84 \times 84$. In our method, we use two coordinate systems, the environment coordinate system (ECS) and the screen coordinate system (SCS). The ECS aims to describe the interaction environment while the SCS aims to visualize and convert to the state space. It should be understood that in the practical environments, this problem is a typical GIS problem since it involves space and pedestrians as the moving objects in case of emergency (Zheni et al. 2009, Xu and Güting 2013, Tryfona et al. 2003). Due to the fact that the inputs are greyscale images (one channel with the value ranging from [0, 255]), the difference between the walls and other moving agents is not distinguished. We set the grayscale images as follows:
1) The background of the images is set to be a zero-value matrix with row $h$ and column $w$, where $h$ and $w$ are also regarded as the height and width of the image.
2) The value for a pixel representing the current pedestrian is set to 255.
3) The value for a pixel representing static obstacles and other pedestrians is set to 100.

#### 3.2.1.2 Action space

In our method, we discretize the action space with eight directions, indexed from 0 to 7. Each agent chooses a direction from the DNN and then translates the direction to a normalized vector $\boldsymbol{v}^{norm}$. The best velocity is calculated as $\boldsymbol{v}^{norm} \times v^{max}$.

#### 3.2.1.3 Reward function

A total of four reward functions are included in this study, including the goal reward function $R^{goal}$, the collision reward function $R^{collision}$, the smooth reward function $R^{smooth}$, and penalty reward function $R^{penalty}$. The total reward R is the aggregation of all four rewards:

$$R = w_1 R^{goal} + w_2 R^{collision} + w_3 R^{smooth} + R^{penalty}, \tag{1}$$

where $w_1$, $w_2$, $w_3$ and $R^{penalty}$ represent the weighting parameters.

$R^{goal}$ defines the reward of one-step movement, described as:

$$R_i^{goal} = \max_j \sum_{j=0}^{N_{exit}} \left(1 - Dist(\boldsymbol{p}_i^t, \boldsymbol{p}_{exit_j})^{w_4}\right) \\ - \max_j \sum_{j=0}^{N_{exit}} \left(1 - Dist(\boldsymbol{p}_i^{t-1}, \boldsymbol{p}_{exit_j})^{w_4}\right) \tag{2}$$

where $Dist(a, b)$ is the Euclidean distance between $a$ and $b$, and $w_4$ is a hyperparameter with the range (0, 1]. The above function consists of two major parts. The first part is the reward based on the current position $\boldsymbol{p}_i^t$ for pedestrian $i$. The second part is the reward based on the last position $\boldsymbol{p}_i^{t-1}$. The $R_i^{goal}$ represents the reward that the current step is relative to the previous one.

$R^{collision}$ defines the reward of the optimal velocity $^{opt}v_i^t$ with the desired velocity towards to exit:

$$R_i^{collsion} = \max_j \sum_{j=0}^{N_{exit}} \frac{Dot(^{opt}v_i^t, Norm(\boldsymbol{p}_{exit_j} - \boldsymbol{p}_i^t))}{v_{max}} \tag{3}$$

where $Norm(\cdot)$ represents the normalization function, $Dot(a, b)$ represents the dot product value, defining the similarity between $a$ and $b$.

$R^{smooth}$ is used to evaluate the smoothness of the pedestrian by comparing the current optimal velocity with the previous optimal velocity:

$$R_i^{smooth} = Dot(^{opt}v_i^t, ^{opt}v_i^{t-1}). \tag{4}$$

### 3.2.2 Rainbow DQN

Reinforcement Learning (RL) generally includes three components, the environment, the agents, and the reward. At each frame t, an agent obtains the state $s_t$ from the environment and conducts an action $a_t$ based on $s_t$. Then the simulation transit to the next state $s_{t+1}$ and the agent receives a reward $r_t$, an evaluation of $r_t$. The interaction continues until the agent reaches the goal, or the iteration number reaches horizon $T$. Theoretically, RL is a branch of the Markov Process Decision (MDP) $< S, A, P, r, \gamma >$, where $S$ is state space, $A$ is action space, $P$ is the transition relationship, $r$ is the reward function, and $\gamma$ is a discount factor that defines the foresight of the agents. An agent's movement satisfies the Markov property, i.e., the current state is only related to the previous state, not to the historical trajectory. The objective of RL is to find a state-action pair mapping that maximizes the total expected return, i.e., $R_t = \sum_{k=0}^{T} \gamma^k r_{t+k}$. In general, a deep neural network (DNN) is required to map the state-to-action function, as the state space and action space tend to be multi-dimensional. In the following part, we briefly introduce the Rainbow DQN algorithm to optimize $R_t$.

Rainbow DQN is an integration of a set of advanced DQN approaches. In our experiments, we utilized Double DQN, DDQN, PER, Multi-Step DQN, Categorical DQN, and Noisy DQN. In DQN, the $Q^\pi(s, a; \theta)$ represents the predicted return in state $s$ with action $a$ following the policy $\pi$ under the parameter setting $\theta$ in the predicted DNN architecture. A bootstrap method (Mnih et al., 2013) is used to represent the target expected return, i.e.,

$$\mathbb{E}[R_t | s_t = s, a_t = a] \approx r_t + \gamma \max_{a_{t+1}} Q(s_{t+1}, a_{t+1}; \hat{\theta}) \quad (5)$$

where $\hat{\theta}$ represents the target DNN parameters, defined to reduce the correlation between predicted expected return and target expected return. To optimize the $R_t$, a Mean Square Error (MSE) loss is defined as

$$L(\theta) = \left( r_t + \gamma \max_{a_{t+1}} Q(s_{t+1}, a_{t+1}; \hat{\theta}) - Q(s, a; \theta) \right)^2 \quad (6)$$

where we only update $\theta$ and $\hat{\theta}$ will be automatically updated to $\theta$ after a constant number of frames ($F_u$).

Double DQN is regarded as a decoupling approach to reduce the overestimation bias, where the target expected return is defined as,

$$\mathbb{E}[R_t | s_t = s, a_t = a] \approx r_t + \gamma Q(s_{t+1}, \underset{a_{t+1}}{\mathrm{argmax}}\, Q(s_{t+1}, a_{t+1}; \hat{\theta}); \theta, \quad (7)$$

DDQN decomposes the last output value, i.e., $Q(s, a; \theta)$ into two streams, the state value $V(s)$ and the advantage value $A(s, a)$, and combines the two streams by a special aggregator to a new output value:

$$Q(s, a; \theta) = V(s) + A(s, a) - \frac{\sum_{a'} A(s, a')}{N_{actions}} \quad (8)$$

where $N_{actions}$ represents a total number of actions. Multi-Step DQN is adopted here to facilitate faster learning. We rewrite the $n$ step reward as:

$$r_t^{(n)} = \sum_{k=0}^{n-1} \gamma_t^{(k)} r_{t+k+1} \quad (9)$$

Thus, the target expected return based on Double DQN can also be rewritten as

$$\mathbb{E}[R_t | s_t = s, a_t = a] \approx r_t^{(n)} + \gamma^{(n)} Q\left(s_{t+1}, \underset{a_{t+1}}{\mathrm{argmax}}\, Q(s_{t+1}, a_{t+1}; \hat{\theta}); \theta\right) \quad (10)$$

To describe the expected value in a more accurate way, Categorical DQN applies the distribution of values, $Z_\theta(s, a)$, instead of a single value, where $\theta$ presents the parameters of the related DNN. $Z_\theta(s, a)$ denotes a discrete distribution, parameterized by $N_{atoms}$. $V_{min}$ and $V_{max}$ respectively denote the minimum of maximum boundaries. For each atom in the distribution $z_i$, defined as $z_i = V_{min} + i\Delta z$, where $i$ is the atom index with the range $[0, N_{atoms})$ and $\Delta z$ is the interval distance between two joint values, calculated as $\Delta z = \frac{V_{max} - V_{min}}{N_{atoms} - 1}$. The predicted return expectation $Q(s, a) = \sum_i z_i p_i(s, a; \theta)$, where $p_i$ is the probability of atom $i$ in state-action. The probability at each atom $i$ is updated as

$$\hat{p}_i = \sum_{j=0}^{N_{atoms}-1} \left[ 1 - \frac{\left| [\hat{z}_j]_{V_{min}}^{V_{max}} - z_i \right|}{\Delta z} \right]_0^1 p_j(s_{t+1}, \underset{a_{t+1}}{\mathrm{argmax}}\, Q(s_{t+1}, a_{t+1}; \hat{\theta}) \quad (11)$$

where $[\cdot]_a^b$ represents the bound of argument in the range $[a, b]$ and $\hat{z}_j$ represents the estimated value for atom $j$, equaling to:

$$\hat{z}_j = r_t + \gamma z_j \tag{12}$$

The KL divergence described as $D_{KL}(\hat{p}||p)$, is applied to evaluate the difference between $\hat{p}$ and $p$ and the cross-entropy term is considered to be the loss function $L(\theta)$ as

$$L(\theta) = -\sum_i \hat{p}_i \log p_i(s, a; \theta) \tag{13}$$

Finally, Noisy DQN is applied to balance the tradeoff between exploration and exploitation. A general linear layer with input $x$ and output $y$ can be represented by

$$y = wx + b \tag{14}$$

where $w$ and $b$ denote the weights and bias respectively. In Noisy DQN, $w$ and $b$ will be re-defined by a Gaussian distribution. Therefore, the related noisy linear layer can be described as

$$y = (\mu^w + \sigma^w \odot \varepsilon^w)x + \mu^b + \sigma^b \odot \varepsilon^b \tag{15}$$

where $\odot$ represents element-wise multiplication to increase the noises in Gaussian distribution. $\mu^w$ and $\sigma^w$ are the parameters in Gaussian distribution parameters for wights $w$. $\mu^b$ and $\sigma^b$ are the parameters for the Gaussian distribution parameters for bias $b$.

To speed up learning efficiency, we applied PER, a sampling-based optimization algorithm. For each data pair $<s_t, a_t, r_t, d_t, s_{t+1}>$, where $d_t$ is a Boolean value to judge whether the agent reaches the destination at time $t$. The corresponding priority $p$ is positively related to $L(\theta)$ as

$$p \propto L(\theta) \tag{16}$$

### 3.2.3 Network architecture

As mentioned above, a state space is composed of the last three greyscale images with a size of $84 \times 84$ and the action space is a discrete space that includes eight different directions, i.e., $N_{actions} = 8$. Based on this setting, the input and output spaces of the DNN are respectively set to be $3 \times 84 \times 84$ and $51 \times 8$. As shown in Figure 2, three two-dimensional convolutional layers are applied to the input $S_t$. After each convolution operation, Batch Normalization (BN) is further applied to prevent overfitting (Ioffe and Szegedy 2015). The output from the convolutional layers is flattened and decomposed as two components. The first component consists of two fully connected (FC) layer with 512 rectifier units and $N_{atoms} \times N_{actions}$ rectifier units, respectively. The second component consists of two fully connected layers with 512 rectifier units and $N_{actions}$ rectifier units, respectively. Noisy terms are added to all the FC layers to encourage exploration at each interaction. Finally, we aggregate the outputs of the two components as the final output, including $N_{actions}$ values for each action in the action space. The activation function in both convolutional and FC layers are the ReLU nonlinearities (Nair and Hinton 2010).

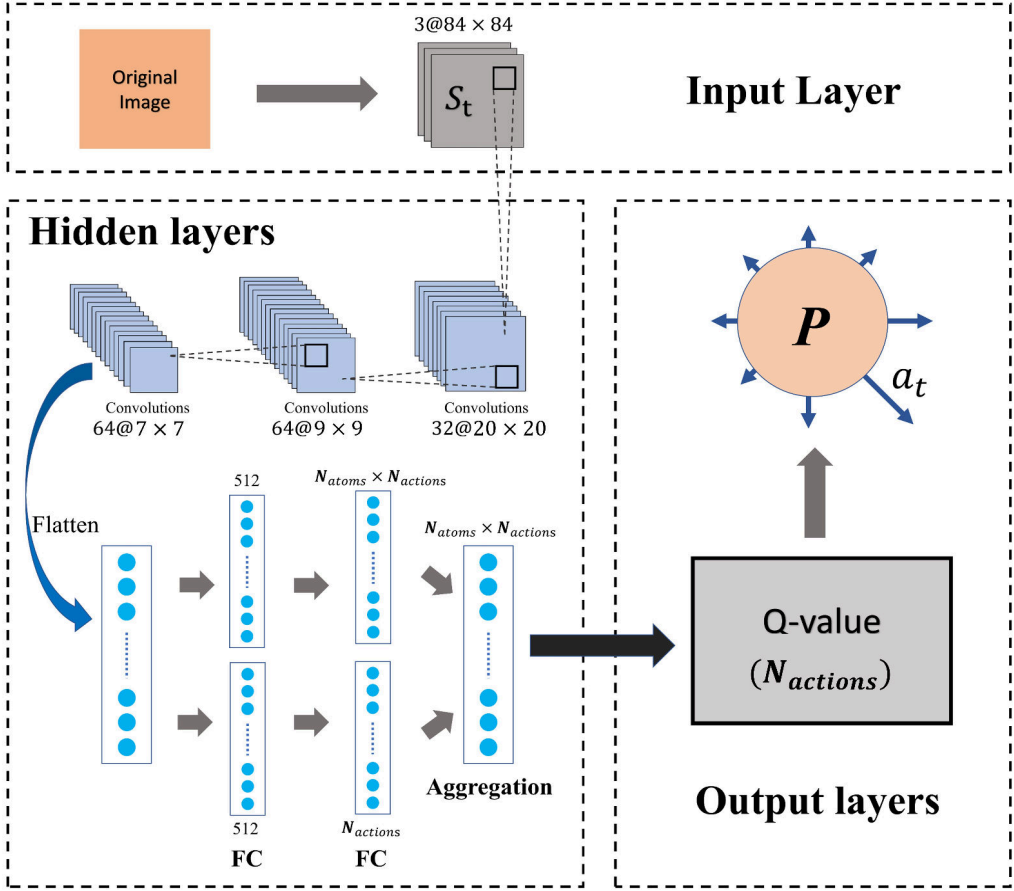

**Figure 2** Network architecture.

## 4. Experiment environment and scenarios

### 4.1 Coding environment

The algorithm of this model is implemented using the Python programming language. PyTorch packages are applied to build a deep neural network (DNN) for mapping the relationship between the state and action. In addition, OpenCV packages are used to collect and visualize data. The program runs on a computer with Ubuntu 18.04 in an environment that consists of i7 CPU, 64G RAM, and two NVIDIA GTX 1080 Ti. The simulation process runs on CPU while the DNN is trained on GPUs. The hyperparameters setting used in this study can be found in Table 1.

**Table 1.** Rainbow DQN hyperparameters used in Multi-Exit evacuation simulation

| Hyperparameter | Value |
| --- | --- |
| Learning rate ($\eta$) | 1e-4 |
| Discount ($\gamma$) | 0.99 |
| Horizon ($T$) | 200 |
| Multi-step ($n$) | 3 |
| Batch size | 128 |
| $F_u$ | 1000 |
| $N_{buffer}$ | 1e+5 |
| $N_{atoms}$ | 51 |
| $V_{max}$ | 10 |
| $V_{min}$ | -10 |
| $w_1$ | 15.0 |
| $w_2$ | 1.25 |
| $w_3$ | 0.5 |
| $w_4$ | 0.4 |
| $R^{penalty}$ | 2.5 |
| $L_s$ | 5e+4 |
| $m$ | 5e+6 |

### 4.2 Scenarios

A virtual indoor environment with two exits named $Exit_l$ and $Exit_b$ respectively, is designed as the research environment. The room geometry is square with a side length of 100 and a wall width of 2.0. The radius of each pedestrian ($r$) equals to 2.0. A total of three scenarios are presented to evaluate the performance of our method. Those scenarios include 1) varying exit width ratio ($p_{ew}$) with a uniform pedestrian distribution, 2) varying pedestrian distribution ratio ($p_{pd}$) with a uniform exit width, and 3) varying exit opening times with a uniform exit width and a uniform pedestrian distribution. In the first scenario, the distribution of the pedestrians and the width of $Exit_b$ ($w_b$) hold the same ($4.0 \times r$) while the width of $Exit_l$ ($w_l$) is set to be $4.0 \times r$, $6.0 \times r$, and $8.0 \times r$, respectively. Assuming $p_{ew}$ represents the ratio of the two exits, i.e., $\frac{w_b}{w_l}$, the three sub scenarios include: $p_{ew} = 1:1$, $p_{ew} = 1:1.5$, and $p_{ew} = 1:2$. This scenario aims to investigate the model's performance under different exit width ratio with uniform distribution of pedestrians. In the second scenario, both exits are with the same width, equaling to $4.0 \times r$. A total of $m$ pedestrians are within the room but with different distributions. Assuming $p_{pd} = \frac{N_{Exit_l}}{N_{Exit_b}}$, where $N_{Exit_l}$ represents the number of pedestrians whose closest exit is $Exit_l$ while $N_{Exit_b}$ represents the number of pedestrians whose closest exit is $Exit_b$, three different distributions are designed, namely $p_{pd} = 1:1$, $p_{pd} = 1:2$ and $p_{pd} = 1:3$. This scenario investigates the model's performance in handling uneven congestions at the exits during the evacuation. In the third scenario, the distribution of pedestrians and the width of the two exits hold the same. However, $Exit_l$ doesn't open at the initial frame while $Exit_b$ keeps open throughout the entire simulation process. In this scenario, the open time for $Exit_l$ is set at the 15[th] frame, the 30[th] frame, and the 45[th] frame, respectively. This scenario creates a dynamic simulation environment as the pedestrians are unaware of when another exit is open. To compare the effectiveness and efficiency of our proposed method, different numbers of pedestrians ($m$) are tested, i.e., $m = 12$, $m = 24$, and $m = 36$ in all three scenarios. We investigate the performance of methods from two perspectives: 1) the total frames for evacuation, and 2) the utilization efficiency of two exits ($r_{util}$). Since the $N_l$ and $N_b$ represent the number of pedestrians to evacuate from $Exit_l$ and $Exit_b$, then the $r_{util}$ can be calculated as:

$$r_{util} = \frac{\min(\frac{N_l}{w_l}, \frac{N_b}{w_b})}{\max(\frac{N_l}{w_l}, \frac{N_b}{w_b})} \qquad (17)$$

With a range of [0,1], $r_{util}$ represents how efficient $Exit_l$ and $Exit_b$ are utilized in general. The higher the $r_{util}$, the more the efficiency of those two exits for the pedestrians to pass during the evacuation. We only investigate $r_{util}$ in the first two scenarios due to the delay imposed to open the $Exit_l$ in the last scenario. The proposed MultiExit-DRL method is compared against AMFM and NLDM from Zheng et al. (2015) and Guo et al. (2012), respectively.

## 5. Results

To compare the performances of this model, we obtain screenshots of different frames during the evacuation process from all three scenarios (Figure 3, Figure 4, and Figure 5). For the first two scenarios, an interval of 10 frames is used, while an interval of 15 frames is used for the third scenario given its longer simulation process. The total frames for pedestrians to evacuate the room and the utilization efficiency of the two exits are analyzed for the three designed scenarios.

### 5.1 Model performances under different exit width ratio ($p_{ew}$)

We first evaluate the model performance under different $p_{ew}$ with uniform distribution of pedestrians. Given different exit width, pedestrians are expected to find appropriate exits to prevent unnecessary congestions, thus leading to high exit utilization efficiency and low total frames (total time for all the pedestrians to evacuate). As expected, the total frames increase along with the increasing number of pedestrians with the same $p_{ew}$ (Table 2). It is reasonable that, with the same indoor environment, it takes a longer time for more pedestrians to evacuate the room. Compared with the other two methods, our proposed MultiExit-DRL method achieves the best performance in all conditions (Table 2). As the number of pedestrians increases (e.g., in the 36-pedestrian case), a larger performance gap is found between MultiExit-DRL and the other two methods, suggesting that MultiExit-DRL can better handle environments with massive agents. For example, in the 36-pedestrian case with $p_{ew} = 1$, the evacuation via AMFM and NLDM takes 115 and 106 frames, respectively, while the evacuation via MultiExit-DRL only takes 60 frames. Since we hold $w_b$ consistently while incrementing the $w_l$, a faster evacuation is achieved by MultiExit-DRL since the wider exit allows more pedestrians to evacuate. This phenomenon proves that, after training with the DRL algorithm, pedestrians are able to recognize the wider exit (the $Exit_l$ in this case) and use it evacuate faster.

**Table 2.** Total frames for evacuation under different exit width ratio ($p_{ew}$) with 12, 24, and 36 pedestrians.

| Methods | 12 pedestrians $p_{ew} =$ | | | 24 pedestrians $p_{ew} =$ | | | 36 pedestrians $p_{ew} =$ | | |
|---|---|---|---|---|---|---|---|---|---|
| | 1:1 | 1:1.5 | 1:2 | 1:1 | 1:1.5 | 1:2 | 1:1 | 1:1.5 | 1:2 |
| AMFM | 51 | 48 | 46 | 69 | 65 | 69 | 115 | 78 | 62 |
| NLDM | 53 | 41 | 47 | 90 | 59 | 70 | 106 | 88 | 54 |
| MultiExit-DRL | **38** | **39** | **34** | **50** | **45** | **45** | **60** | **57** | **52** |

*Note.* Best performances among the three methods are highlighted in bold.

**Table 3.** The utilization efficiency ($r_{util}$) under different exit width ratio ($p_{ew}$) with 12, 24, and 36 pedestrians.

| Methods | 12 pedestrians $p_{ew} =$ | | | 24 pedestrians $p_{ew} =$ | | | 36 pedestrians $p_{ew} =$ | | |
|---|---|---|---|---|---|---|---|---|---|
| | 1:1 | 1:1.5 | 1:2 | 1:1 | 1:1.5 | 1:2 | 1:1 | 1:1.5 | 1:2 |
| AMFM | 0.71 | 0.25 | 0.18 | 0.71 | 0.90 | 0.29 | **0.89** | 0.36 | 0.06 |
| NLDM | **1.00** | **0.93** | 0.67 | **1.00** | 0.75 | 0.29 | **0.89** | 0.50 | 0.18 |
| MultiExit-DRL | **1.00** | **0.93** | **0.70** | 0.85 | **0.93** | **0.70** | 0.80 | **0.83** | **1.00** |

Note. Best performances among the three methods are highlighted in bold.

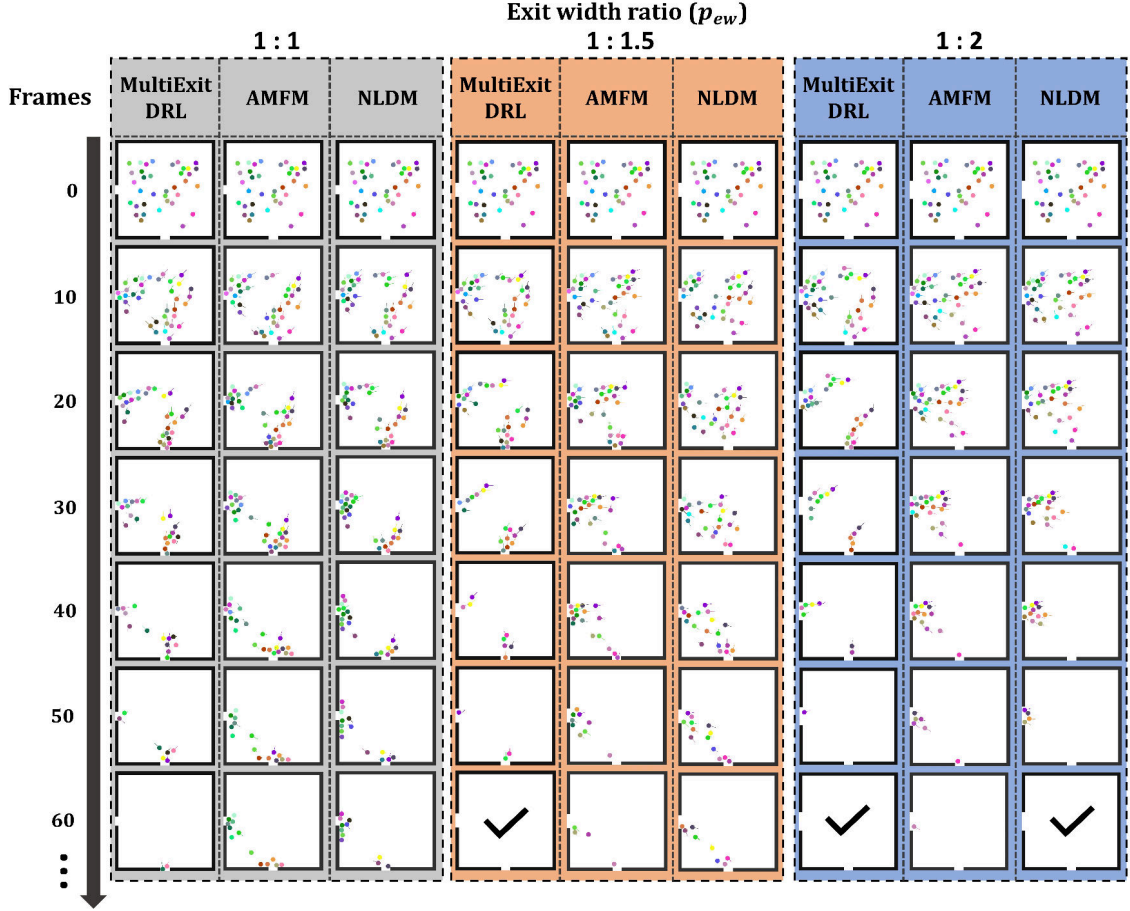

**Figure 3** Model performance for 36 pedestrians under different exit width ratios.

The efficiency of exit utilization of this scenario is presented in Table 3. All the methods achieve decent performances with $p_{ew} = 1:1$ (i.e., $w_b = w_l$), suggesting that all three models can handle the situation where the two exits have the same width, creating the same attraction for all the pedestrians. However, the superiority of MultiExit-DRL is clear when the two exits become unbalanced. For example, in the 36-pedestrian case with $p_{ew} = 1:2$ (i.e., $w_l = 2 \times w_b$), the values of $p_{ew}$ in AMFM and NLDM are 0.06 and 0.18, respectively while the value of $p_{ew}$ in MultiExit-DRL is 1.00 (Table 3), suggesting that the efficiency of exit utilization in MultiExit-DRL is significantly better with an unbalanced exit width ratio compared with the other two methods. The superiority of MultiExit-DRL is well documented from the screenshots of frames during the evacuation process (Figure 3). When $p_{ew} = 1:2$, pedestrians in AMFM and NLDM are clearly crowded at $Exit_l$, evidenced by the screenshots of the 20[th], the 30[th], and the 40[th] frame (Figure 3). Even though the width of $Exit_l$ is twice larger than the width of $Exit_b$, $Exit_l$ lacks the ability to handle overwhelming pedestrians. This congestion is significantly due to the low efficiency of exit utilization with many frames of the AMFM and the NLDM method. In comparison, pedestrians in MultiExit-DRL are able to choose exits more appropriately, given the unbalanced $p_{ew}$, leading to the perfect efficiency of exit utilization with few total frames.

**5.2 Model performances under different pedestrian distribution ratio ($p_{pd}$)**

This scenario investigates the model's performance in handling different initial distributions of pedestrians. Given the different distribution patterns (i.e., a different $p_{pd}$), pedestrians are expected to adjust their strategies during the evacuation to exit the room as fast as possible with high efficiency of exit utilization. As shown in Table 4, our proposed MultiExit-DRL method achieves the best performance in all of the designed conditions. Similar to the comparison in the first scenario, MultiExit-DRL significantly outperforms the other two methods as the number of pedestrians increases regardless of the variations of the initial distributions, thus demonstrating its great capability in handling massive agents. In addition, the insensitivity to the different initial distributions in our method suggests that pedestrians have learned to adjust their strategies appropriately to prevent congestions at different initial locations. For instance, in a crowded 36-pedestrian case, pedestrians in MultiExit-DRL evacuate the room in 60 and 58 frames for $p_{pd}$ = 1:1 and $p_{pd}$ = 1:3, respectively (Table 4). Despite the fact that $p_{pd}$ = 1:1 suggests an even distribution while $p_{pd}$ = 1:3 suggests the number of pedestrians whose initial locations are closer to $Exit_b$ is three times as many as $Exit_l$, both rooms are evacuated within a similar number of frames (Table 4).

**Table 4.** Total frames for evacuation under different pedestrian distribution ratio ($p_{pd}$) with 12, 24, and 36 pedestrians.

| Methods | 12 pedestrians $p_{pd}$ = | | | 24 pedestrians $p_{pd}$ = | | | 36 pedestrians $p_{pd}$ = | | |
|---|---|---|---|---|---|---|---|---|---|
|  | 1:1 | 1:2 | 1:3 | 1:1 | 1:2 | 1:3 | 1:1 | 1:2 | 1:3 |
| AMFM | 51 | 53 | 58 | 69 | 103 | 121 | 115 | 117 | 119 |
| NLDM | 53 | 66 | 65 | 90 | 83 | 104 | 106 | 149 | 153 |
| MultiExit-DRL | **38** | **51** | **44** | **50** | **56** | **53** | **60** | **61** | **58** |

*Note.* Best performances among the three methods are highlighted in bold.

**Table 5.** The utilization efficiency ($r_{util}$) under different pedestrian distribution ratio ($p_{pd}$) with 12, 24, and 36 pedestrians.

| Methods | 12 pedestrians $p_{pd}$ = | | | 24 pedestrians $p_{pd}$ = | | | 36 pedestrians $p_{pd}$ = | | |
|---|---|---|---|---|---|---|---|---|---|
|  | 1:1 | 1:2 | 1:3 | 1:1 | 1:2 | 1:3 | 1:1 | 1:2 | 1:3 |
| AMFM | 0.71 | **0.71** | **0.33** | 0.71 | 0.41 | 0.33 | **0.89** | 0.57 | 0.57 |
| NLDM | 1.00 | **0.71** | **0.33** | **1.00** | 0.71 | 0.71 | **0.89** | 0.71 | **0.89** |
| MultiExit-DRL | **1.00** | 0.33 | **0.33** | 0.85 | **1.00** | **0.85** | 0.80 | **0.80** | **0.89** |

*Note.* Best performances among the three methods are highlighted in bold.

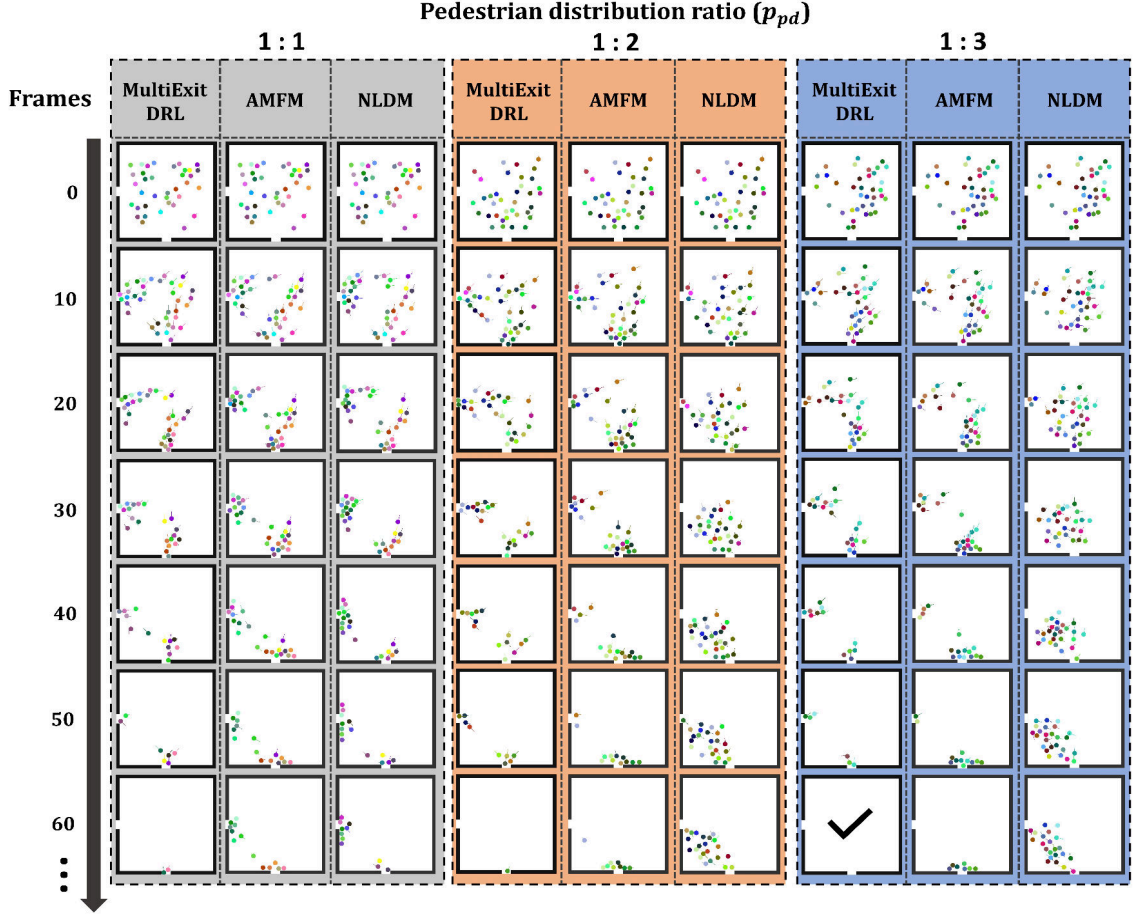

**Figure 4** Model performance for 36 pedestrians under different pedestrian distribution ratios.

In terms of exit utilization, higher $r_{util}$ values are found for MultiExit-DRL in the crowded environments (cases with 24 and 36 pedestrians), where all $r_{util}$ values are above 0.8, suggesting great efficiency of exit utilization (Table 5). In the 12-pedestrian case, however, MuiltiExit-DRL presents low $r_{util}$ with uneven initial distributions ($p_{pd} = 1:2$ and $p_{pd} = 1:3$). It indicates that, in the current hyperparameter setting, pedestrians in MultiExit-DRL prioritize closer exit in the uncrowded environment. Despite the low $r_{util}$, pedestrians in MultiExit-DRL can still evacuate the room faster than the other two methods, as evidenced by the low number of the total frames (Table 4). It is observed that the initial uneven distribution might cause congestion at the exit for certain methods, which is well documented by the screenshots of frames. When $p_{pd} = 1:2$ and $p_{pd} = 1:3$, pedestrians in AMFM clearly congest at $Exit_b$, as $Exit_b$ is the closest exit to the initial locations of most pedestrians. However, this congestion unavoidably results in low efficiency of exit utilization and longer evacuation time. In comparison, pedestrians in MultiExit-DRL can clearly adjust their evacuation strategies, leading to a balanced exit assignment. The $30^{th}$ frame in all cases shows that, with our proposed MultiExit-DRL method, both exits are targeted with a balanced amount of pedestrians, which largely increase the efficiency of exit utilization and reduces the evacuation time.

### 5.3 Model performances under different opening times of $Exit_l$

The different opening times of a certain exit creates a dynamic simulation environment where pedestrians are expected to recognize the newly opened exit and adjust their evacuation strategies accordingly. Similar to the previous two scenarios, MultiExit-DRL shows the least total frames for pedestrians to evacuate the room compared with AMFM and NLDM in all conditions (Table 6). As the number of pedestrians increases, the superiority of MultiExit-DRL becomes obvious. In addition, the earlier the $Exit_l$ opens, the faster the pedestrians in MultiExit-DRL evacuate, especially in a more crowded environment, e.g., the 24- and 36-pedestrian case. Since $Exit_b$ is the only available exit before opening $Exit_l$, all pedestrians are moving in one direction towards $Exit_b$ (Figure 5). The discrepancy of pedestrians' behaviors occurs when pedestrians start to be aware of the availability of $Exit_l$. In the case when $Exit_l$ opens at the 15$^{th}$ frame, about half of the pedestrians in MultiExit-DRL switched to the new exit, thus to greatly speed up the evacuation process. In comparison, pedestrians in AMFM and NLDM usually fail to respond timely, and if they respond, they cause congestion due to lack of the unbalanced loads (see AMFM pedestrians in the 45$^{th}$ frame when $Exit_l$ opens at the 15$^{th}$ frame).

To sum up, three different scenarios are designed to compare the performance of our proposed method, MultiExit-DRL against AMFM and NLDM. The model's performance is revealed from the total number of frames and the efficiency of exit utilization, i.e., the $r_{util}$. The results indicate remarkable superiority of MultiExit-DRL in terms of total frames, as pedestrians under the MultiExit-DRL method are able to evacuate the room with the least frames in all designed conditions. With the room becoming more crowded by adding additional pedestrians, the performance gap becomes more obvious, suggesting great generalization capability of the MultiExit-DRL method. As for the efficiency of exit utilization, MultiExit-DRL shows consistent high $r_{util}$ values in the first scenario with a varying $p_{ew}$, indicating that pedestrians under MultiExit-DRL can adjust their strategies according to different exit widths once the evacuation phase starts. MultiExit-DRL also exhibits high $r_{util}$ values in crowded environments in the second scenario with a varying $p_{pd}$. Although it presents low $r_{util}$ in the 12-pedestrian case, the evacuation is still completed within fewer frames compared with AMFM and NLDM. The underperformance of AMFM can be explained by its intrinsic design. As an improved adaptive multi-factor model, it utilizes an additional judgment to prevent pedestrians from frequently changing the targeted exit. Despite its great performance in uncrowded environments, pedestrians in AMFM fail to modify their strategies timely when more pedestrians are added to the environment. Therefore, congestion at one exit and idleness at the other is found in AMFM, especially in crowded environments. NLDM, as a nested logit discrete model, uses non-strict logical judgment, potentially leading to frequent changes of targeted exits of its pedestrians, consequently leading to more frames. Different from the two methods above, MultiExit-DRL provides a proper direction for each pedestrian under the current state via Rainbow DQN, allowing the pedestrians to choose the optimal directions of movement by given reward functions. It further allows pedestrians to rapidly explore other options instead of congesting at a certain exit.

**Table 6.** Total frames for evacuation under different open frames for $Exit_l$ with 12, 24, and 36 pedestrians.

| Methods | 12 pedestrians | | | 24 pedestrians | | | 36 pedestrians | | |
|---|---|---|---|---|---|---|---|---|---|
| | $Exit_l$ opens at frame | | | $Exit_l$ opens at frame | | | $Exit_l$ opens at frame | | |
| | 15$^{th}$ | 30$^{th}$ | 45$^{th}$ | 15$^{th}$ | 30$^{th}$ | 45$^{th}$ | 15$^{th}$ | 30$^{th}$ | 45$^{th}$ |
| AMFM | 66 | 74 | 68 | 100 | 106 | 136 | 144 | 159 | 149 |
| NLDM | 70 | 68 | 68 | 97 | 141 | 183 | 153 | 183 | 173 |
| MultiExit-DRL | **48** | **49** | **49** | **74** | **86** | **100** | **84** | **104** | **121** |

*Note.* Best performances among the three methods are highlighted in bold.

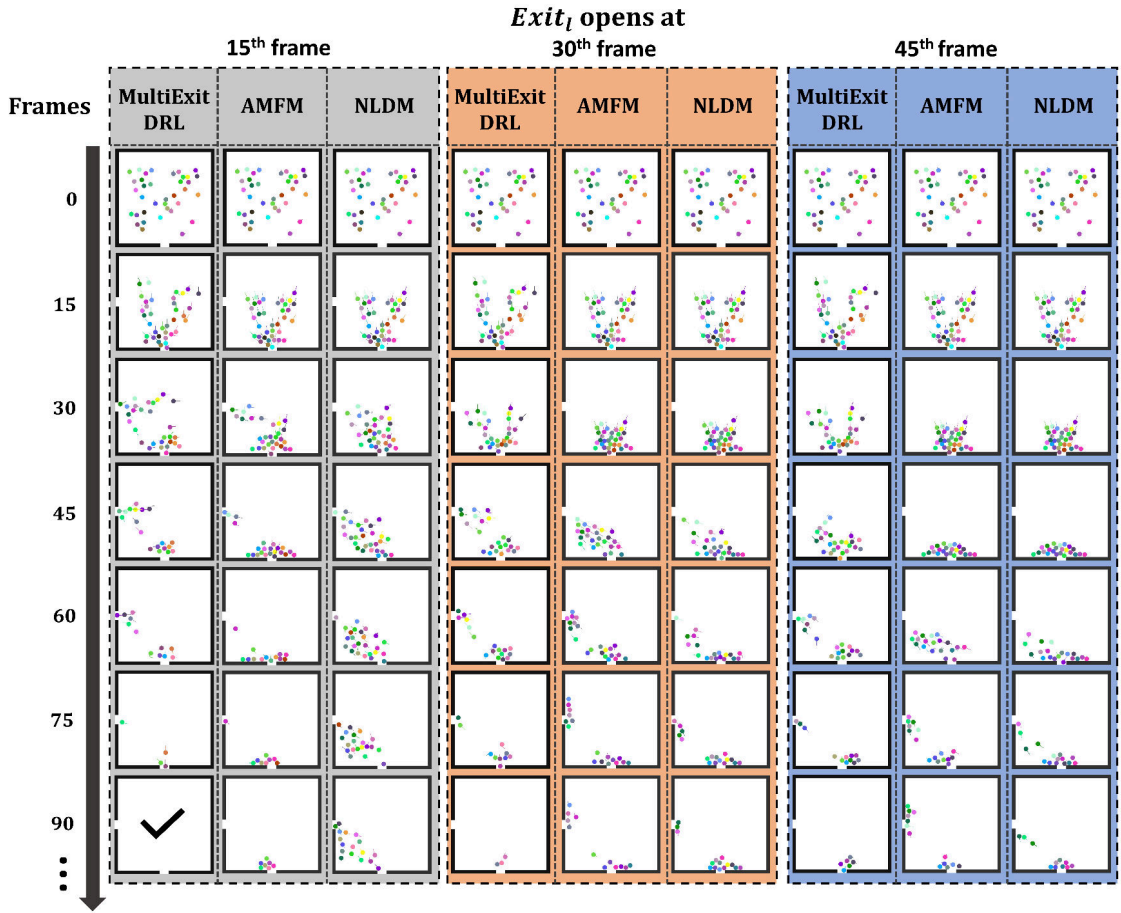

**Figure 5** Model performance for 36 pedestrians under different opening times of $Exit_l$.

## 6. Discussion

In this study, we presented a novel multi-exit evacuation simulation by integrating ORCA and DRL. In general, our method shows great performances both in the training process and the evacuating process. Specifically, it outperforms other approaches due to its 1) efficient learning speed, 2) stability, 3) ability to scale up and 4) great exit utilization.

Transcending popular ray-based state acquisition approaches that require intensive computational resources (Lee et al. 2018, Xu et al. 2020), we directly capture screenshots of the environment via OpenCV as the source of information, allowing the pedestrians to learn their surroundings quickly. Besides, by packing adjacent frames as input to the DNN, no external information (such as position and velocity) is required. Given the exponential growth of the complexity in the ray-based methods when the number of pedestrians increases, the complexity of our method stays with linear growth that makes it more efficient to perform simulations with massive pedestrians. Stability is another merit of our approach. In the multi-exit simulation, pedestrians often hesitate when facing multiple exits, given the limited designs of traditional mathematical models. Although researches suggested that this problem can be mitigated by a proper probability assignment (Zheng et al. 2017), still the existing models fail to present stable results. In our method, pedestrians are granted a global perspective by the application of DRL for selecting the optimal action in the current state. Given the long-term reward maximum nature of DRL, pedestrians keep interacting with the environment for numerous times to learn the best action for the total return. Once the best state-to-action mapping function has been learned by the pedestrians, they take actions without hesitation, allowing the simulation process to be pretty stable. Our method is tested with a different number of pedestrians, and the results have shown great performance compared with other traditional methods. This scaling-up ability is due to the fact that the effectiveness of our approach largely depends on the reward function rather than the complexity of the environment. As the number of pedestrians increases, the state space expands accordingly, which in turn leads to longer training time. However, the increase of pedestrians does not affect pedestrians taking optimal action to maximize the value of the reward function in the given state. Finally, the results proved a great exit utilization as pedestrians in our approach hardly crowd at a certain exit. For traditional multi-exit selection methods, once a pedestrian chooses an exit at a certain frame, it moves towards the target exit, regardless of any changes, thus causing congestion that eventually reduces the evacuation efficiency. In our approach, however, instead of choosing a specific exit for a pedestrian in a given frame, an optimal direction is calculated based on DRL. This optimal direction prevents further crowding at a certain exit by allowing pedestrians to explore other opportunities. This behavior can be achieved via the implementation of an appropriate reward function via DRL but is difficult to be expressed via traditional multi-exit selection simulation that utilizes mathematical models.

Despite the merits explained above, it worth noting that our model is designed under the following constraints. Limited by the discrete nature of Rainbow DQN, we adopted a discrete space with a total of eight possible directions. It turns out that, in a few cases, pedestrians move along with Manhattan distances instead of Euclidean distances with better efficiency. The application of DRL algorithms with a continuous action space like DDPG, PPO, or SAC might further improve the model performance. The total reward function in our method is composed of four sub-reward functions, i.e., the goal reward function, the collision reward function, the smooth reward function, and the penalty reward function, where each sub-reward function needs a hyperparameter to specify its importance. However, the balance between the four hyperparameters requires trial-and-error experiments which are not sufficiently covered in our work.

Future studies should focus on designing more simple and reliable reward functions. It is acknowledged that the performances of DRL algorithms greatly rely on the number of training samples. Our approach uses a single-core data collection method that basically causes sub-optimal simulations under the same experimental configuration and reward function. This can be regarded as the result of incomplete exploration of agents. In the future, the potentially more advanced DRL methods, such as the APeX-DQN (Horgan et al. 2018), need to be explored. Finally, the pedestrians in our method have the same size and velocity. However, in practice, each pedestrian is not uniform. Our future research will extend this method to accommodate heterogeneous pedestrians.

## 7. Conclusion

Multi-exit evacuation simulation is one of the key areas that need more attention due to its potential in public safety. Traditional multi-exit evacuation simulation methods largely rely on discrete multi-exit selection methods that are characterized by different factors such as distance, density, and exit width. These methods are, however, coupled with low exit utilization and congestion at the exits. In this article, we proposed a novel multi-exit evacuation simulation that integrates ORCA and DRL, referred to as the MultiExit-DRL, where local collision avoidance detection is achieved via ORCA, and movement direction is achieved via DRL. We further designed a DNN framework to facilitate state-to-action mapping. In the designed framework, successive screenshots (greyscale images) are used as the raw state, and they have proven to be faster in data collection compared to ray-based state acquisition. The action space is further divided into eight isometric directions for pedestrians to vacate. Rainbow DQN, a DRL algorithm that integrates several advanced DQN methods, is applied to improve data utilization and algorithm stability. We compared our proposed MultiExit-DRL method with two traditional multi-exit evacuation simulation models, i.e., the AMFM and the NLDM in three individual scenarios: 1) varying pedestrian distributions with a uniform exit width; 2) varying exit widths with a uniform pedestrian distribution; 3) varying exit opening schedules with a uniform exit width and a uniform pedestrian distribution. The results have shown that the proposed MultiExit-DRL presents great learning efficiency in all of the designed experiments. It further shows great utilization of exits regardless of the number of pedestrians. Nevertheless, MultiExit-DRL has some limitations, such as the utilization of discrete action space and homogeneous pedestrian design. Further researches should focus on solving these problems and investigate the model's performance in large areas with more complicated scenarios.

**Data availability statement**

No third-party data was used in this study. The source code and videos for demonstration can be found on GitHub at https://github.com/XD1227/MultiExit-Rainbow